\def\year{2022}\relax
\documentclass[letterpaper]{article} 
\usepackage{aaai22}  
\usepackage{times}  
\usepackage{helvet}  
\usepackage{courier}  
\usepackage[hyphens]{url}  
\usepackage{graphicx} 
\usepackage{natbib}  
\usepackage{caption} 
\DeclareCaptionStyle{ruled}{labelfont=normalfont,labelsep=colon,strut=off} 
\usepackage{algorithm}
\usepackage{algorithmicx}
\usepackage{algpseudocode}

\usepackage{amsmath}
\usepackage{amsthm}
\usepackage{booktabs}
\usepackage{epsfig}
\usepackage{multirow}
\usepackage{subfigure}
\usepackage{setspace}
\usepackage{amssymb}
\usepackage{color}

\usepackage{newfloat}
\usepackage{listings}

\newcommand{\etal}{\textit{et al.}}
\newcommand{\eg}{\textit{e.g.}}
\newcommand{\etc}{\textit{etc.}}
\newcommand{\ie}{\textit{i.e.}}

\floatstyle{ruled}
\newfloat{listing}{tb}{lst}{}
\floatname{listing}{Listing}
\title{PMAL: Open Set Recognition via Robust Prototype Mining}
\author{
	Jing Lu\textsuperscript{\rm 1}\equalcontrib, 
	Yunlu Xu\textsuperscript{\rm 1}\equalcontrib, 
	Hao Li\textsuperscript{\rm 1}, 
	Zhanzhan Cheng\textsuperscript{\rm 1,2}\thanks{Corresponding author.},
	Yi Niu\textsuperscript{\rm 1}
}
\affiliations{
	\textsuperscript{\rm 1} Hikvision Research Institution, Hangzhou, China\\
	\textsuperscript{\rm 2} Zhejiang University, Hangzhou, China\\
	\{lujing6, xuyunlu, lihao50, chengzhanzhan, niuyi\}@hikvision.com
}

\begin{document}

\maketitle

\begin{abstract}
	Open Set Recognition (OSR) has been an emerging topic. 
	Besides recognizing predefined classes, the system needs to reject the unknowns.
	Prototype learning is a potential manner to handle the problem,
	as its ability to improve intra-class compactness of representations is much needed in discrimination between the known and the unknowns.
	In this work, we propose a novel Prototype Mining And Learning (PMAL) framework.
	It has a prototype mining mechanism before the phase of optimizing embedding space,
	explicitly considering two crucial properties, namely \textit{high-quality} and \textit{diversity} of the prototype set.
	Concretely, a set of high-quality candidates are firstly extracted from training samples
	based on data uncertainty learning, avoiding the interference from unexpected noise.
	Considering the multifarious appearance of objects even in a single category, a diversity-based strategy for prototype set filtering is proposed.
	Accordingly, the embedding space can be better optimized to discriminate therein the predefined classes and between known and unknowns.
	Extensive experiments verify the two good characteristics (\textit{i.e.}, \textit{high-quality} and \textit{diversity}) embraced in prototype mining, and show the remarkable performance of the proposed framework compared to state-of-the-arts.
\end{abstract}

\section{Introduction} \label{intro}
Classic image classification problem is commonly based on the assumption of \textit{close set}, \ie, categories appeared in testing set should all be covered by training set.
However, in real-world applications, samples of unseen classes may appear in testing phase, which will inevitably be misclassified into the specific known classes.
To break the limitations of \textit{close set}, Open Set Recognition (OSR) \cite{Scheirer2013Toward} was proposed, which has two sub-goals: known class classification and unknown class detection.

\begin{figure}[t]
	\begin{center}
		\includegraphics[width=1.\linewidth]{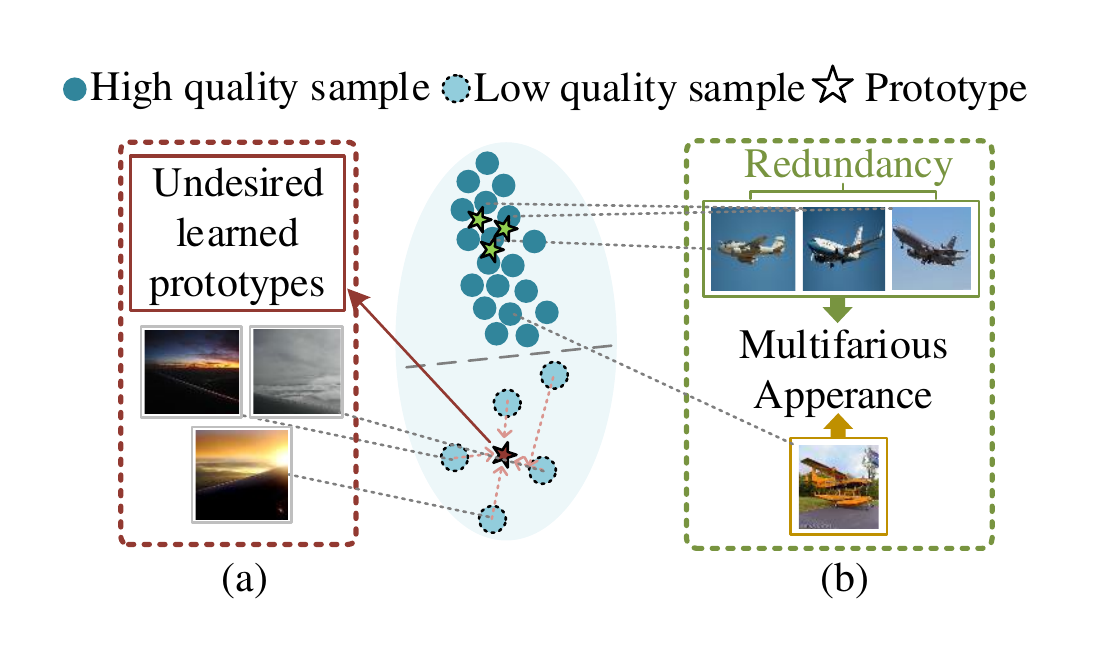}
	\end{center}
	\caption{
		Two typical problems on {implicitly} learned prototypes. (a) {Undesired learned prototypes arisen from low-quality samples.} (b) Redundancy in similar prototypes and lacks of diversity. The in-between ellipse shows the feature distribution of an exemplary class `airplane'.
	}
	\label{fig-instabiity}
\end{figure}

Methods based on \textit{Prototype Learning} (PL) obtained promising performance \cite{Yang2018Robust,Chen2020Learning} recently.
This group generates clearer boundaries between the known and unknowns through learning more compact intra-class feature representations using \textit{prototypes} (on behalf of the discriminative features of each class).
In detail, \cite{Yang2018Robust} learns the CNN feature extractor and prototypes jointly from the raw data and predicts the categories by finding the nearest prototypes instead of the traditional SoftMax layer.
\cite{Chen2020Learning} advanced the framework \cite{Yang2020Convolutional} by adversely using the prototypes named reciprocal points to represent the outer embedding space of each known class, and then limiting the embedding space of unknown class. 
{
	The existing methods all conduct prototype learning and embedding optimization jointly,
	regarding the prototypes as parameterized vectors, 
	without direct constraint on the procedure of obtaining prototypes.
	Here we call them \textit{implicitly} learned prototypes, 
	and oppositely, if imposing direct guidance on the prototypes themselves, 
	we denote the prototypes as \textit{explicitly} learned ones.
	All the above-mentioned methods belong to the former category, \ie, the \textit{implicit} prototype-based methods.
	While they inevitable encounter some problems, especially in complicated situations.}
Two typical problems are shown in Figure \ref{fig-instabiity}:
{
	(1) \textbf{Undesired learned prototypes close to feature space of low-quality\footnote[1]{Low quality can be caused by various noise, \eg, occlusion, blur or background interference.} samples}. 
	As in Figure \ref{fig-instabiity}(a), implicitly learned prototypes are mistakenly guided by the low-quality samples.
	As claimed in \cite{Shi2019Probabilistic},
	embedding of \textit{high-quality} samples is discriminative while low-quality samples correspond to {ambiguous features}. 
	Prototypes should represent the discriminative features of each class,
	so only \textit{high-quality} samples are suitable.
}
(2) \textbf{Redundancy in similar prototypes and lack of diversity.}
Without explicit guidance, prototypes in one category show much redundant and cannot sufficiently represent the multifarious appearance.
As in Figure \ref{intro}(b), prototypes marked in green are adjacent in feature space and samples nearby show similar appearance, which implies the redundancy in learned prototypes. Besides,
the airplanes in green and yellow rectangles show great distinctions, and their embeddings are located at separated positions.
Obviously only using prototypes in green can not fully captures the multifarious appearance, which we require the \textit{diversity} of prototype set.

\begin{figure}
	\begin{center}
		\includegraphics[width=1.0\linewidth]{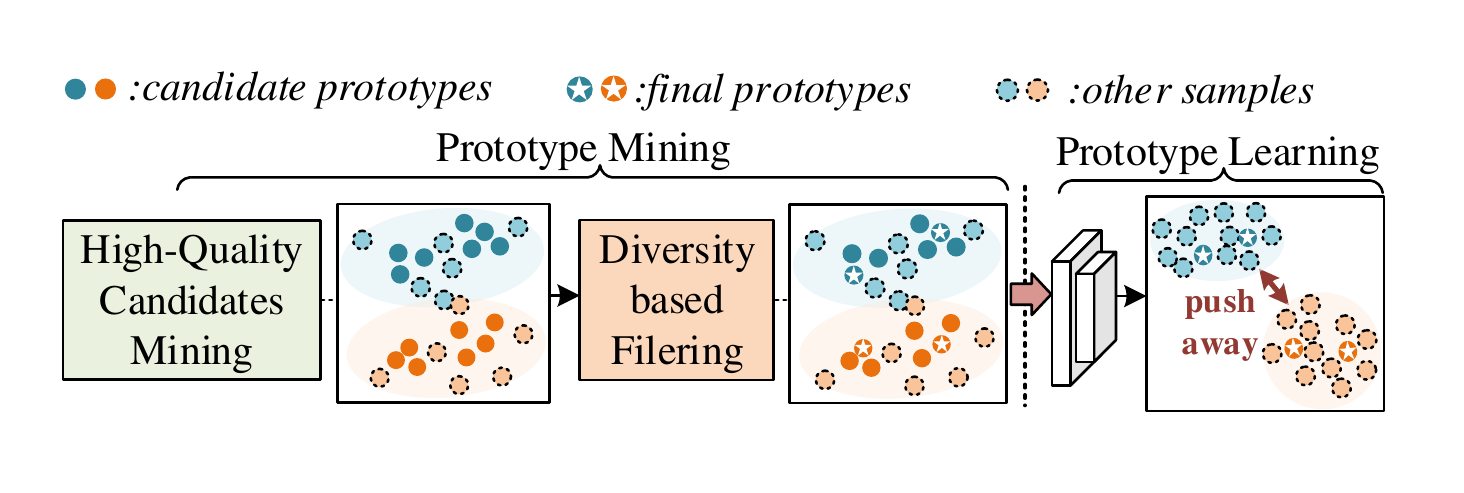}
	\end{center}
	\caption{{The proposed prototype mining and learning framework. Different colors denote different classes.}}
	\label{fig-architecture}
\end{figure}
Upon the above, we take \textit{high-quality} prototypes and their \textit{diversity} into consideration
and propose to explicitly design the prototype\footnote[2]{In our method, prototypes refer to samples, not features.} mining criteria, and then conduct PL with the chosen desirable prototypes.
Note that different from the existing \textit{implicit} prototypes, 
our proposed can be regarded as \textit{explicit} ones.
We name the novel framework as Prototype Mining And Learning (PMAL).
The framework is illustrated in Figure \ref{fig-architecture},
which can be divided into two phases, the prototype mining and embedding learning orderly.
(1) \textbf{The prototype mining phase}.
\textit{High-quality} candidates are first extracted from training set
according to the novelly proposed metric \textit{embedding topology robustness}, {which captures the data uncertainty contained in samples arisen from inherent low-quality factors}.
Then the prototype set filtering is designed to incorporate \textit{diversity} for prototypes in each class.
The step not only prevents the redundancy of similar prototypes,
but also preserves the multifarious appearance of each category.
(2) \textbf{The embedding optimization phase.} In this phase, given high-reliable prototypes,
the embedding space is optimized via a well-designed point-to-set distance metric. 
{
	The training burden is also reduced via mining prototypes in advance and feature optimization orderly,
	as the latter phase only work on embedding space.}


Our main contributions are as follows.
{
	(1) Different from the common usage of implicitly learnable prototypes, we pay more attention on choosing prototypes with explicit criteria for OSR tasks. We point out the two important attributes of prototypes, namely the \textit{high-quality} and \textit{diversity}.
}
(2) 
We design a OSR framework by prototype mining and learning.
In the \textit{prototype mining} phase, the above two key attributes are taken into consideration.
In the \textit{embedding learning} phase, with the chosen prototypes as fixed anchors for each class, {a better embedding space is learned, without any sophisticated skills for convergence}.
(3) Extensive experiments on multiple OSR benchmarks show that our method is powerful to discriminate the known and unknowns, surpassing the state-of-the-art performance by a large margin,
especially in complicated large-scale tasks.

\section{Related work}
\label{related_work}
OSR is theoretically defined by Scheirer \etal \cite{Scheirer2013Toward}, where they added an hyperplane to distinguish unknown samples from knowns in an SVM-based model.
With rapid development of deep neural networks, Bendale \etal \cite{Bendale2016Towards} incorporated deep neural networks into OSR by introducing the OpenMax function. 
Then both Ge \cite{Ge2017Generative} and Neal \cite{Neal2018Open} tried to synthesize training samples of unseen classes via the popular Generative Adversarial Network.

Recently, reconstruction-based \cite{Yoshihashi2019Classification,Poojan2019C2AE,Sun2020Conditional} approaches are widely studied, among which
Sun \etal \cite{Sun2020Conditional} achieved promising results by learning conditional gaussian distributions for known classes then detecting unknowns.
Zhang \etal \cite{Zhang2020Hybrid} added a flow density estimator on top of existing classifier to reject unseen samples.
These methods all incorporate auxiliary models (\eg, auto-encoders)  for OSR, thus inevitably bring extra computational cost.

Since \cite{Yang2020Convolutional,Chen2020Learning} attempted to combine prototype learning with deep neural networks for OSR, they achieved the new state-of-the art.
Prototypes refer to representative samples or latent features for each class.
It is inspired by \textit{Prototype Formation} theory in psychology cognition field \cite{Rosch1973Natural}, and is later incorporated in some deep networks, \eg, face recognition \cite{Ma2013Prototype,Wen2016Prototype}, few-shot learning \cite{Snell2017Prototypical}.
Yang \etal \cite{Yang2020Convolutional,Yang2018Robust} introduced Convolutional Prototype Network (CPN), in which prototypes per class were jointly learned during training. 
Chen \etal \cite{Chen2020Learning}  learned discriminative reciprocal points for OSR, which can be regarded as the inverse concept of prototypes.
{
	However, these methods suffer from unreliable prototypes caused by low-quality samples and lack of diversity, leading to the limited representativeness of prototypes.}
\section{Notation and Preliminaries}
\subsection{Notations}
Let $X$$\to$$Z$ denote the mapping from input dataset $X$=$\{x_i\}_{i=1}^N$ into its embedding space $Z$=$\{z(x_i)\}_{i=1}^N$ by a trained deep classification model, where $Z$$\in$$\mathbb{R}^{N\times D}$, $N$ is the number of samples and $D$ is the embedding channel size.
{
	The feature region occupied by samples of the known class $k$ in $Z$ is referred as \textit{embedding region} $Z_k$ where $k$$\in$$\{1,...,K\}$, $K$ is the number of known classes.}

Given the input $x_i$$\in$X, the extracted feature $z(x_i)$ (simply denoted as $z_{i}$) is fed into the ultimate linear layer, then SoftMax operation is conducted to obtain the probability $p(\cdot)$ of $x_i$ belonging to the $k$-th class, which is:
\begin{small}
	\begin{equation}\label{cls}
		p(\hat{y}_{i}=k|z_{i})=\mathit{exp}({z_{i}{w_{k}}+b_k})/\sum_{n=1}^{K}{\mathit{exp}({z_{i}{w_n}+b_n})},
	\end{equation}
\end{small}
where $\hat{y}_{i}$ is predicted class and $W\text{=}(w_1,...,w_K)\in\mathbb{R}^{D \times K}$ and $b\in$$\mathbb{R}^{K}$ are the weight and bias term of the linear layer. 

\subsection{Preliminaries of Uncertainty}\label{DU}
In deep uncertainty learning, \textit{uncertainties} \cite{Chang2020Data} can be categorised into \textit{model uncertainty} and \textit{data uncertainty}.
\textit{Model uncertainty} captures the noise of parameters in deep neural networks.
What we mention in this work is \textit{data uncertainty}, which captures the inherent noise in input data.
It has been widely explored in deep learning to tackle various computer vision tasks, \eg, face recognition \cite{Shi2019Probabilistic}, semantic segmentation \cite{Alex2015Bayesian} \etc. 
{
	Generally, inherent noise is attributed to two factors: the low quality of image and the label noise.
	In the scope of this work for assessing qualified samples in PL, we only regard the former. 
	Following \cite{Chang2020Data}, when mapping an input sample $x_i$ into $Z$,
	its inherent noise, \ie, \textit{data uncertainty}, contained in input will also be projected into embedding space, the embedded feature $z(x_i)$ can be formulated as:
	\begin{small}
		\begin{equation}\label{def}
			z(x_i)=\phi(x_i)+n(x_i), \ n(x_i)\sim\mathcal{N}(0,\sigma(x_i))
		\end{equation}
	\end{small}
	where {$\phi(x_i)$ represents the discriminative class-relevant feature of $x_i$}, 
	{which can be seen as the ideal embedding for representing its identity}.
	$\phi$ denotes the embedding model. 
	$n(x_i)$ is drawn from a Gaussian distribution with mean of zero and $x_i$-dependent variance $\sigma(x_i)$,
	$\sigma(x_i)$ represents the \textit{data uncertainty} (\textit{i.e.}, class-irrelevant noisy information caused by low quality) of $x_i$ in $Z$.
	\textit{The more noise contained in $x_i$, the larger uncertainty $\sigma(x_i)$ exists in embedding space.}}
We denote $z(x_i)$, $\phi(x_i)$, $\sigma(x_i)$ as $z_{i}$, $\phi_{i}$, $\sigma_{i}$ for simplicity hereinafter.

\section{Prototype Mining}
Prototype mining phase has two steps orderly, the \textit{high-quality} candidate selection and \textit{diversity}-based filtering.
\subsection{High-Quality Candidate Selection}
{

	Since data uncertainty captures noise in samples caused by low-quality, we exploit it for selecting high-quality samples as candidate prototypes. To model data uncertainty, a simple yet efficient algorithm is proposed, which includes the following three steps:
	1) embedding space initialization, 2) data uncertainty modeling and 3) candidate selection.
	
	\subsubsection{Embedding Space Initialization.}	
	Following Monte-Carlo simulation\cite{Yarin2016Dropout}, we first acquire $U$ SoftMax-based deep classifiers $\{M^{u}\}_{u=1}^{U}$ on the training set of known classes by repeating the training process $U$ times.
	Then the input data is fed into the pre-trained classifiers, obtaining $\{Z^{u}\}_{u=1}^{U}$.
	Noticing that it is sufficient to formalize different embedding space by conducting repeated training processes with random parameter initialization and data shuffling, as proved in \cite{Balaji2017Simple}. Here we set $U$ to 2 for clearer illustration.
	
	\subsubsection{Data Uncertainty Modeling.}
	Based on Sec. 3.2, the higher quality for a sample, the lower data uncertainty it has.
	
	\textbf{Property 1.} \textit{Given a high-quality sample $x_i$, its embedding $z_{i}$ satisfies $z_{i}$$\approx$$\phi_{i}$.}
	
	The high-quality sample $x_i$ satisfy $\sigma_{i}$$\approx$$0$, then combined with Equa. \ref{def} we can easily obtain the above property.
	Suppose we select high-quality samples from training data to form the candidate prototype set $C$=$\{c_i\}_{i=1}^{H}$$\subseteq$$X$, where $H$ is the total candidate number.
	Correspondingly, the set of their embedding in two different space $Z^1$ and $Z^2$ can be denoted as $ \Phi^1$=$ \{z_{i}^1\}_{i=1}^H$$\approx$$ \{\phi_{i}^1\}_{i=1}^{H}$ and $ \Phi^2$=$\{z_{i}^2\}_{i=1}^H$$\approx$$ \{\phi_{i}^2\}_{i=1}^{H}$, where the superscript denotes the index of embedding space.
	
	\textbf{Property 2.} \textit{Given a sample pair $(x_i,x_j)$, $\forall$ $i,j$ $\in$$\{1,...H\}$, Mahalanobis distance in embedding space $Z$ can be computed by $ d_{\mathcal{M}}(z_{i},z_{j})\text{=} \sqrt{(z_{i}\text{-}z_{j})\varSigma^{-1}(z_{i}\text{-}z_{j})^{\mathrm{T}}}$ where $\varSigma^{-1}$ is covariance matrix.
		If $x_i$, $x_j$ are both of high quality, $ d_{\mathcal{M}}(z_{i},z_{j})$ in different embedding space remains similar,
		\ie,
		$ d_{\mathcal{M}}(z_{i}^1,z_{j}^1)$$ \approx$$ d_{\mathcal{M}}(z_{i}^2,z_{j}^2)$, $\forall x_i,x_j \in C$.
	}
	
	\textbf{\textit{Proofs.}}
	When only feeding the class-relevant feature $\phi_{i}^1$ and $\phi_{i}^2$ into the top linear layer of each classifier, the output probability for each category should remain consistent under the constraint of same class label $y_{i}$, \ie, $p(\hat{y}_{i}^{1}\text{=} k|\phi_{i}^{1})$$\approx$$p(\hat{y}_{i}^{2}\text{=} k|\phi_{i}^{2})$, $\forall$$k$$\in$$\{1,...,K\}$.
	Combining Equa. \ref{cls}, we have the formulation:
	\begin{small}
		\begin{equation}\label{equa-etr-derivation}
			\frac{\mathit{exp}({\phi_{i}^1{w_k^1}+b_k^1})}{\sum_{n=1}^{K}{\mathit{exp}({\phi_{i}^1{w_n^1}+b_n^1}})} \approx  \frac{\mathit{exp}({\phi_{i}^2{w_k^1}+b_k^1})}{\sum_{n=1}^{K}{\mathit{exp}({\phi_{i}^2{w_n^1}+b_n^2})}}, \\
		\end{equation}
	\end{small}
	which can be deduced to
	\begin{small}
		\begin{equation}\label{equa-etr-derivation}
			\begin{aligned}
				\phi_{i}^1(w_n^1\text{-}w_k^1)\text{+}b_n^1\text{-}b_k^1 &\approx \phi_{i}^2(w_n^2\text{-}w_k^2)\text{+}b_n^2\text{-}b_k^2, \ n\text{=}1,...,K.
			\end{aligned}
		\end{equation}
	\end{small}
	Averaging up all the equations for $\forall$ $k$$\in$$\{1,...,K\}$ leads to
	\begin{small}
		\begin{equation}\label{equa-etr-derivation2}
			\begin{aligned}      \phi_{i}^1(w_n^1\text{-}\overline{w}^1)\text{+}b_n^1\text{-}\overline{b}^1 \approx \phi_{i}^2(w_n^2\text{-}\overline{w}^2)\text{+}b_n^2\text{-}\overline{b}^2,\ n\text{=}1,...,K,
			\end{aligned}
		\end{equation}
	\end{small}
	where $ \overline{w}$=($ \sum_{l=1}^{K}{w_l}$)/$ K$ and $\overline{b}$=($ \sum_{l=1}^{K}{b_l}$)/$ K$.
	Taking $A$= $(w_1-\overline{w},...,w_K-\overline{w})$ and $B$=$(b_1-\overline{b},...,b_K-\overline{b})$, Equa. \ref{equa-etr-derivation2} can be rewritten as $ \phi_{i}^1A^1$+$ B^1 \approx \phi_{i}^2A^2$+$ B^2$.
	Given another $ c_{j}$$ \in$$ C$ where $j$$\neq$$i$, the same equation $ \phi_{j}^1A^1$+$ B^1 \approx \phi_{j}^2A^2$+$ B^2$ can be obtained. Combining these two equations leads to $(\phi_{i}^1 - \phi_{j}^1)A^1 \approx (\phi_{i}^2 - \phi_{j}^2)A^2$, which is equivalent to:
	\begin{small}
		\begin{equation}\label{equa-etr-derivation}
			\begin{aligned}
				\sqrt{(\phi_{i}^1 \text{-} \phi_{j}^1)A^1{A^1}^\mathrm{T}(\phi_{i}^1 \text{-} \phi_{j}^1)^{\mathrm{T}}}\approx\sqrt{(\phi_{i}^2 \text{-} \phi_{j}^2)A^2{A^2}^\mathrm{T}(\phi_{i}^2 \text{-} \phi_{j}^2)^{\mathrm{T}}}
			\end{aligned}
		\end{equation}
	\end{small}

	Here, $AA^{\mathrm{T}}\text{=}(w_1-\overline{w},...,w_K-\overline{w})(w_1-\overline{w},...,w_K-\overline{w})^{\mathrm{T}}$.
	As \cite{Chang2020Data} pointed out, $w_n$$\in$$\{w_1,...w_K\}$ in $A$ can be seen as the center (or mean) of embedding region $Z_n$, \ie, $E(z_{i}|y_{i}\text{=}n)$$\approx$$w_n$. 
	Thus $AA^{\mathrm{T}}$ is a reasonable estimation for the covariance matrix $\varSigma$$^{-1}$ of $Z$.
	Consequently, Equa. \ref{equa-etr-derivation} educes $ d_{\mathcal{M}}(z_{i}^1,z_{j}^1)$$ \approx$$ d_{\mathcal{M}}(\phi_{i}^1,\phi_{j}^1)$$ \approx$$ d_{\mathcal{M}}(\phi_{i}^2,\phi_{j}^2)$$ \approx$$ d_{\mathcal{M}}(z_{i}^2,z_{j}^2)$.
	
	
	\textbf{Definition 1. \textit{Embedding Topology Robustness.}} \textit{Given a sample $x_i$, its relative position to other samples in embedding space $Z$ is defined by `embedding topology' as: $t({z_{i}}) \triangleq (d_\mathcal{M}(z_{i},z_{1}),...,d_\mathcal{M}(z_{i},z_{{N}}))$.
		Then the distance metric `embedding topology robustness' is defined by:}
	\begin{small}
		\begin{equation}\label{equa-robustness}
			r(x_i)\triangleq \mathit{exp}({-||t(z_{i}^1)-t(z_{i}^2)||_2})
		\end{equation}
	\end{small}
	where $||$$\cdot$$||_2$ is Euclidean distance. Following \textit{Property 2}, $r(\cdot)$ possesses the following characteristic.
	
	\textbf{Property 3.} \textit{High-quality samples have large \textit{embedding topology robustness} $r(\cdot)$ near 1, while low-quality ones correspond to smaller $r(\cdot)$.}
	
	
	\begin{figure}
		\begin{center}
			\includegraphics[width=1.\linewidth]{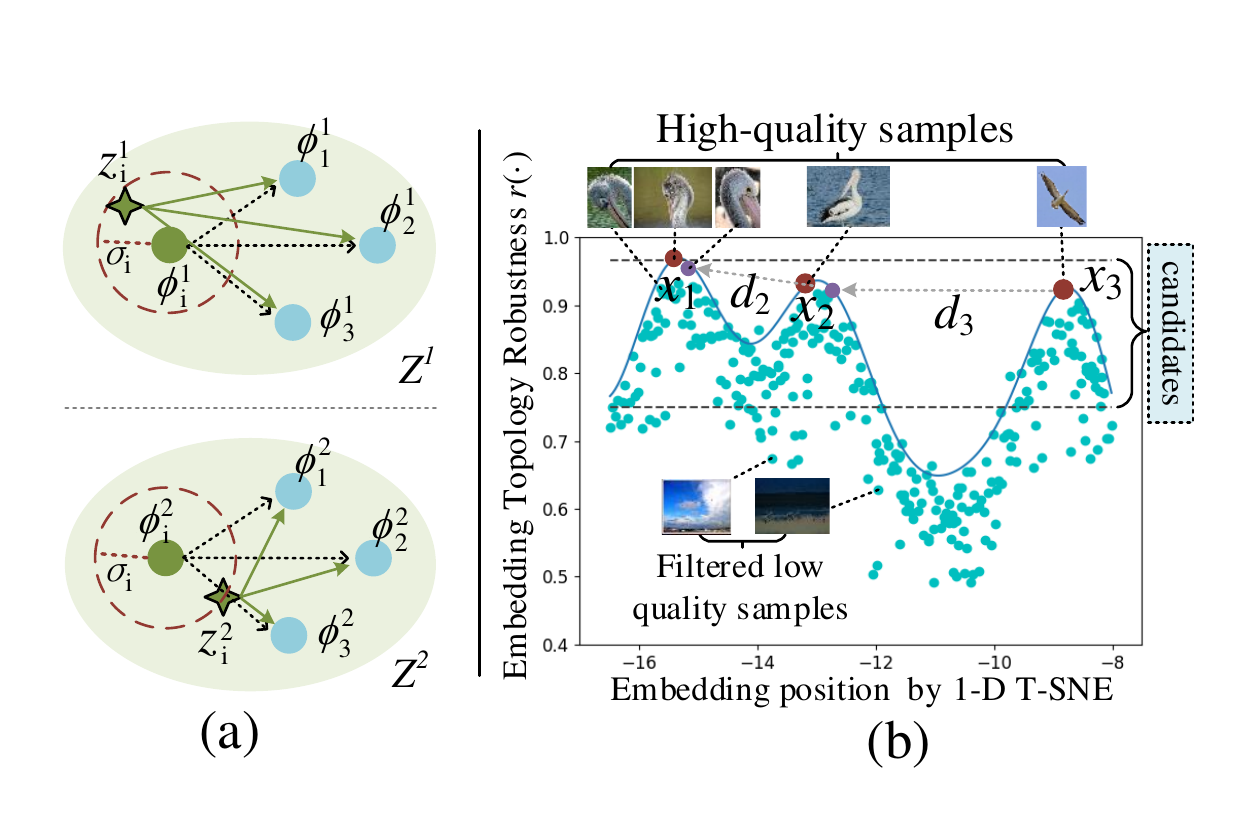}
		\end{center}
		\caption{(a) Illustration for the effect of data uncertainty $\sigma_{o_i}$ on \textit{Embedding Topology Robustness}.
			(b)The Distribution of $r(\cdot)$ for class `Pelican' on ImageNet. The blue curve fits the upper contour of the distribution.}
		\label{fig-method}
	\end{figure}
	For a high-quality sample $x_i$$\in$$C$ with data uncertainty $\sigma_{i}$$\approx$0, since $d_{\mathcal{M}}(z_{i}^1,z_{j}^1)$$\approx$$d_{\mathcal{M}}(z_{i}^2,z_{j}^2)$, $\forall x_j$$\in$$C$, then $||t({z_{i}^1})- t({z_{i}^2})||_2$ will be a small value approaching 0, hence robustness $r(x_i)$ will be a large value near 1.
	
	For a low-quality sample $x_i$$\in$($X$$\setminus$$C$) with large uncertainty $\sigma_{i}$, the consistency of \textit{Embedding Topology} will be disrupted. See Figure \ref{fig-method}(a), the Mahalanobis distance from class-relevant feature $\phi_{i}$ to $\phi_{1}$, $\phi_{2}$, $\phi_{3}$ remains similar in $Z^1$ and $Z^2$ following above analysis, thus the topology shape among $\phi_{(\cdot)}$ (dashed arrows) keeps unchanged. But $z_{{i}}^1$ and $z_{{i}}^2$ vary evidently caused by $\sigma_{{i}}$, hence topology shape from $z_{i}$ to $\phi_{1}$, $\phi_{2}$, $\phi_{3}$ (green solid arrows) shows great distinctions in two space,
	which results in a reduced $r(x_{i})$.
	Obviously, the larger uncertainty $\sigma_{i}$ will trigger larger variation of topology shape, leading to smaller $r(x_i)$.
	\subsubsection{Candidate Selection.}
	We denote the set of all input $x_{i}$ in class $k$ as $\mathrm{S}_{k}$.
	To generate candidate prototype set $C_k$ for class $k$,
	we first find the sample with the highest \textit{embedding topology robustness} score, \ie, $\max \{ r(x_{i})|x_{i}\in\mathrm{S}_{k} \}$. 
	Then samples with $r(\cdot)$ value above $\epsilon \cdot \max \{ r(x_{i})|x_{i}\in\mathrm{S}_{k} \}$ are elected into $C_k$, where $\epsilon$ is a preset threshold.
}

\subsection{Diverse Prototype-Set Filtering}
After selecting all the \textit{high-quality} images into the candidate set $C$=$\{C_i\}_{i=1}^K$, two problems await:
(1) $C$ can be \textit{highly redundant}. As in Figure \ref{fig-method}(b), samples near $x_1$ share similar appearance and features. Such redundancy will bring extra computation cost in the next multi-prototype learning step. A straightforward way is to design a filtering for removing the redundant;
(2) The multifarious visual appearance of object within the same class leads to distinguished feature representations. For example in Figure \ref{fig-method}(b), $x_1$, $x_2$ and $x_3$ appear in different visual looking and their embedding are located at separated positions,
symbolizing the diversity of embedding.
Such diversity of embedding should be preserved during filtering. 

Upon above, the task is turned to generate final prototype set $P$=$\{P_k\}_{k=1}^K$ from the obtained candidate set $C$ considering both \textit{high-quality} and \textit{diversity}.
Specifically for each class $k$, the method should find samples with local maximum $r(\cdot)$ and large embedding distance to form $P_k$, like the $x_1$, $x_2$ and $x_3$ in Figure \ref{fig-method}(b). 

Similar to the NP-hard \textit{coreset selection} \cite{Ozan2018Active} problem, our goal is to choose $T$ prototypes from $C_k$ into $P_k$ for each class $k$. We implement it by iteratively collecting prototypes by a greedy algorithm as
\begin{small}
	\begin{equation}\label{div_obj}
		P_{k}\text{=}\bigcup_{i=1}^{T}\{x_{i}|{\max\limits_{x_i \in C_{k}} \{\min\limits_{x_j \in C_{k}} d_\mathcal{M}(z_{i},z_{j})|r(x_j)>r(x_i)\}}\}.
	\end{equation}
\end{small}
For initialization, we search candidates with the max $r(\cdot)$ in $C_k$ through $\max \{ r(x_{i})|x_{i}\in\mathrm{C}_{k} \}$  to initialize $P_k$, then append candidates satisfying Equation \ref{div_obj} into $P_k$ in an iterative way.
The detailed implementation is given in Algorithm \ref{algo-diversity}.

\begin{algorithm}[h]
	\caption{Filter Candidate Prototype Set with Diversity}
	\label{algo-diversity}
	\begin{algorithmic}[1]
		\Require
		Candidate prototype set $C$=$\{C_i\}_{i=1}^K$;
		Class number $K$;
		Prototype number per class $T$;
		\Ensure
		final prototype set $P$=$\{P_k\}_{k=1}^K$;
		\For{$k=1$ to $K$}
		\State compute Mahalanobis distance matrix $D_k$$\in$$\mathbb{R}^{{N_k}\text{x}{N_k}}$
		\Statex \qquad in $Z^1$ (or $Z^2$), $N_k$ is the candidate number in $C_k$; 
		\State initial a $N_k$-length array $E$ with max value in $D_k$;
		\For{$i=1$ to $N_k$}
		\State for $i$-th candidate $x_{i}$$\in$$C_k$, find its closest can-
		\Statex \qquad \quad didate $x_{j}$ in $C_k$, where $r(x_{j})$$>$$r(x_{i})$, if exists,
		\Statex \qquad \quad update $E$[i]=$D_k$[i,j];
		\EndFor
		\State sort $E$ in descending order, $E_{ind}$ is the sorted index
		\Statex \qquad array;
		\Repeat
		\State add sample whose index is $E_{ind}$[0] in $C_k$ into
		\Statex \qquad \quad $P_k$, then remove $E_{ind}$[0] from $E_{ind}$;
		\Until{the number of samples in $P_k$ exceeds $T$}
		\EndFor
	\end{algorithmic}
\end{algorithm}
Taking Figure \ref{fig-method}(b) for example, $x_1$ has the max $r(\cdot)$ thus is first elected, then $x_3$ and $x_2$ are successively added into final set, as they correspond to 2$^{\text{nd}}$/3$^{\text{rd}}$ largest value $d_3$ and $d_2$ in $E$.
Note that Mahalanobis distance of high-quality samples remains similar in $Z^1$ or $Z^2$, thus computing $D_k$ in either space leads to similar selected prototypes.
\section{Embedding Optimization}
Generated prototypes as anchors to represent known classes, we enlarge the distance between different embedding regions to reserve larger space for unknowns. Thus the risk of unknowns misclassified as known classes can be reduced.
\subsection{Prototype-based Space Optimization}
Given sample $x_i$ belonging to known class $m$ and $P_k$=$\{p_{k,l}\}_{l=1}^{T}$, we denote the distance from $x_i$ to prototype set $P_k$ as $d(z_{i}, z({P_k}))$, where $z({P_k})$$=$
$(z({p_{k,1}}),...,z({p_{k,T}}))$$\in$$\mathbb{R}^{D\times T}$ is the embedding of $T$ prototypes in $P_k$. Then we incorporate a prototype-based constraint to optimize a better embedding space for OSR:
\begin{small}
	\begin{equation}\label{equa-optimize}
		\begin{aligned}
			\mathcal{L}_{p} \text{=}\frac{1}{N}\sum_{i=1}^{N}[&d(z_{i}, z({P_m}))-d(z_{i},z({P_u}))+\delta]_{+}, \\
			\ P_u &\text{=}\mathop{\arg\min}_{P_k \in {P\setminus P_m}}(d(z_{i},z({P_k})))
		\end{aligned}
	\end{equation}
\end{small}
where $P_u$ is the closest prototype set among other classes and $\delta$ is a tunable margin.
Unlike existed methods \cite{Yang2018Robust,Chen2020Learning} that jointly learn sample embedding $z_{i}$ and prototype representation $z({P_k})$ in training, we update $z({P_k})$ by directly feeding fixed prototype samples in $P_k$ into current embedding model, thus our model can focus on learning a better sample embedding $z_{i}$.
{
	Such training strategy is more advantageous since we not only avoid the unstable learning of $z({P_k})$ , but also ease training difficulty of $z_{i}$.}
Finally, the loss in training phase is a combination:
\begin{equation}\label{equa-full}
	\mathcal{L}=\mathcal{L}_{cls}+\lambda_{p}\mathcal{L}_{p},
\end{equation}
where $\mathcal{L}_{cls}$ is the SoftMax loss and $\lambda_{p}$ is a balancing coefficient.
Besides, we design a {new point-to-set distance metric with self-attention \cite{Vaswani2017Attention} mechanism to effectively measure the distance $d(z_{i}, z({P_k}))$}. 
\begin{small}
	\begin{equation}\label{equa-optimize2}
		\begin{aligned}
			&d(z_{i}, z({P_k}))\text{=}1\text{-}\frac{{z_{i}}^{\mathrm{T}}z^{att}_i({P_k})}{|{z_{i}}^{\mathrm{T}}||z^{att}_i({P_k})|}, \\ &z^{att}_i({P_k})\text{=}\textit{SoftMax}(\frac{z_{i}^{\mathrm{T}}z({P_k})}{\sqrt{d}})z({P_k})
		\end{aligned}
	\end{equation}
\end{small}
where $\sqrt{d}$ is a scale factor\cite{Vaswani2017Attention} and $|\cdot|$ denotes L2 norm. We use $z_{i}$ to query embedding in $z({P_k})$ to get its similarity with each prototype, and obtain weighted sum $z^{att}_i({P_k})$.
Then distance is computed by referring to similarity between $z_{i}$ and $z^{att}_i({P_k})$.
We jointly consider the correlations between $x_i$ and all diversified prototypes, thus measure the point-to-set distance more comprehensively.
\subsection{Rejecting Unknowns}
Following the general routine \cite{Yang2020Convolutional}, two rejection rules are adopted for detecting unknown samples: (1) \textit{Probability based Rejection (PR)}. We directly reject unknowns by thresholding SoftMax probability scores;
(2) \textit{Distance based Rejection (DR)}. Unknowns are rejected by thresholding the minimum point-to-set distance, \ie $min\{{d(z_{i},z({P_k}))}\}$ where $P_k$$\in$$P$, since unknown samples should have larger distance with the closest prototype set than known samples.

\section{Experiments}
\subsection{Experiments on Small-Scale Benchmarks}\label{small}
\textbf{Datasets.}
Following \cite{Neal2018Open}, we first conduct comparisons with state-of-the-arts on 6 standard datasets including
(1) MNIST \cite{LeCun1998Gradient}, SVHN \cite{Netzer2011Reading}, CIFAR10 \cite{Krizhevsky2009Learning}: 4 classes are randomly selected as known classes and the rest 6 classes are unknowns; (2) CIFAR+10, CIFAR+50: 4 non-animal classes from CIFAR10 are chosen to be known classes, then 10 and 50 animal classes are respectively sampled from CIFAR100 \cite{Krizhevsky2009Learning} to be unknowns; 
(3) TinyImageNet (TINY) \cite{Le2015Tiny}: 20 classes are randomly sampled as knowns and the left 180 classes are unknowns.

\subsubsection{Implementations.}
Two backbones are adopted to implement our method. The light-weighted backbone OSCRI \cite{Neal2018Open} with parameters less than $1M$ is used to validate our performance when equipped on applications with limited resources. The larger-scale backbone Wide-ResNet (WRN) \cite{Chen2020Learning} with $9M$ parameters is implemented for a fair comparison with previous methods, whose parameter is still less than most existed methods \cite{Yoshihashi2019Classification,Poojan2019C2AE,Sun2020Conditional}.
We adopt Adam optimizer to train our model on each dataset for 600 epochs with batchsize 128.
The learning rate starts at 0.01 and is dropped by 0.1 every 120 epochs, momentum is set to 0.9 and weight decay is 5e-4. The same optimization strategy is used for obtaining pre-trained models and for embedding space optimization. For all datasets, the margin $\delta$ is fixed to 0.5, $\lambda_{p}$ is set to 1 and prototype number $T$=10.

\subsubsection{Evaluation Protocols.}
Following \cite{Neal2018Open}, the evaluation includes 2 parts: (1) close set performance of known classes is reported by classification accuracy ACC on test set of knowns, and (2) unknown detection performance is evaluated by the most adopted metric AUROC (Area Under ROC Curve) \cite{Neal2018Open} on the test set of both known and unknown classes.
Reported results are averaged over 5 random splits. We observe two rejection rules lead to similar results, thus we simply report the results of \textit{DR}.

\begin{table*}[t]
	\caption{Close set ACC and Open set AUROC on small datasets. `*' denotes implemented results and `C' is short for `CIFAR'.}
	\label{sota_small_acc}
	\centering
	\scalebox{0.85}{
		\begin{tabular}{c|cccccc|cccccc}
			\hline
			\multirow{2}{*}{Methods} & \multicolumn{6}{c|}{Close set ACC} & \multicolumn{6}{c}{Open set AUROC} \\
			\cline{2-13}
			\specialrule{0em}{1pt}{1pt}
			& MNIST & SVHN & C10 & C+10 & C+50 & TINY & MNIST & SVHN & C10 & C+10 & C+50 & TINY  \\
			\hline
			SoftMax & {99.5} & 94.7 & 80.1 & - & - & -   & 97.8 & 88.6 & 67.7 &  81.6 & 80.5 & 57.7  \\
			CPN (\citeauthor{Yang2020Convolutional}) & 99.7 & 96.7 & 92.9 & 94.8$^{*}$ & 95.0$^{*}$ & 81.4$^{*}$   & 99.0 & 92.6 & 82.8 & 88.1 & 87.9 & 63.9  \\
			PROSER (\citeauthor{Zhou2021Placeholders}) & - & 96.5 & 92.8 & - & - & 52.1  & 94.3 & - & 89.1 & 96.0 & 95.3 & 69.3 \\
			CGDL (\citeauthor{Sun2020Conditional}) & 99.6 & 94.2 & 91.2 & - & - & -   & 99.4 & 93.5 & 90.3 & 95.9 & 95.0 & 76.2  \\
			OpenHybrid (\citeauthor{Zhang2020Hybrid}) & 94.7 & 92.9 & 86.8 & - & - & -   & 99.5 & 94.7 & 95.0 & 96.2 & 95.5 & 79.3  \\
			RPL-OSCRI (\citeauthor{Chen2020Learning}) & 99.5$^{*}$ & 95.3$^{*}$ & 94.3$^{*}$ & 94.6$^{*}$ & 94.7$^{*}$ & 81.3$^{*}$  & 99.3 & 95.1 & 86.1 & 85.6 & 85.0 & 70.2 \\
			ARPL (\citeauthor{Chen2021Adversarial}) & 99.5 & 94.3 & 87.9 & 94.7 & 92.9 & 65.9  & 99.7 & 96.7 & 91.0 & 97.1 & 95.1 & 78.2 \\
			RPL-WRN (\citeauthor{Chen2020Learning}) & 99.6$^{*}$ & 95.8$^{*}$ & 95.1$^{*}$ & 95.5$^{*}$ & 95.9$^{*}$ & 81.7$^{*}$  & 99.6 & 96.8 & 90.1 & 97.6 & 96.8 & 80.9 \\
			\hline
			PMAL-OSCRI & 99.6 & 96.5 & 96.3 & 96.4 & 96.9 & 84.4  & 99.5 & 96.3 & 94.6 & 96.0 & 94.3 & 81.8 \\
			PMAL-WRN & \textbf{99.8} & \textbf{97.1} & \textbf{97.5} & \textbf{97.8} & \textbf{98.1} & \textbf{84.7} & \textbf{99.7} & \textbf{97.0} & \textbf{95.1} & \textbf{97.8} & \textbf{96.9} & \textbf{83.1} \\
			\hline
		\end{tabular}
	}
\end{table*}

\subsubsection{Result Comparison.}
(1) \textbf{{Close Set Recognition}}:  
Table \ref{sota_small_acc} shows we obtain the best ACC on all datasets, especially the gain reaches 2\%$\sim$3\% on three CIFAR datasets and TINY.
We attribute it to the fact that PMAL learns more compact intra-class embedding compared to other methods (shown in Figure \ref{fig-vis-embed-comparison}(e)$\sim$(h), thus the classification decision boundaries among classes can be more correctly drawn.
(2) \textbf{{Open Set Recognition}}:
PMAL achieves the best AUROC on all benchmarks in Table \ref{sota_small_acc}, especially on the most complex TINY, PMAL-WRN achieves 2.2\% gain compared to previous best RPL-WRN.
The superiority of PMAL is more obvious when equipped on light-weight `OSCRI'. Compared to RPL-OSCRI, ARPL and CPN with the same backbone, PMAL-OSCRI outperforms them by a larger margin over 3.6\%.
Besides,
the light PMAL-OSCRI only falls slightly behind PMAL-WRN, still holding top performance among all the reported results (even those with larger networks). 

\subsection{Experiments on Larger-Scale Benchmarks}
\subsubsection{Datasets.} We further validate our method on more challenging large-scale datasets including
(1) ImageNet-100, ImageNet-200 \cite{Yang2020Convolutional}: 
the first 100 and 200 classes from ImageNet \cite{Deng2009ImageNet} are selected as known classes and the rest are treated as unknowns;
(2) ImageNet-LT \cite{Liu2019Long}: a long-tailed dataset with 1000 known classes from ImageNet-2012 \cite{Deng2009ImageNet}, and additional classes in ImageNet-2010 are as unknowns. The image number per class ranges from 5 to 1280, thus it can well simulate the problem of long-tailed data distribution.
\subsubsection{Implementations.}
Similar to \cite {Chen2020Learning}, ResNet-50 \cite{He2016Deep} is used as network backbone and SGD optimizer is adopted with learning rate 0.2, which drops by 0.1 every 30 epochs. Other detailed setups are the same with experiments on small-scale datasets in Section \ref{small}.
\subsubsection{Result Comparison.}
\begin{table*}[t]
	\caption{Comparisons on 3 large-scale datasets. We denote `ImageNet' as `IN' for simplicity.}
	\label{sota_imagenet}
	\centering
	\scalebox{0.9}{
		\begin{tabular}{c|ccc|ccc|ccc}
			\hline
			\multirow{2}*{Method} & \multicolumn{3}{c|}{Close Set ACC} & \multicolumn{3}{c|}{Open Set AUROC} & \multicolumn{3}{c}{Additional Params}    \\
			\cline{2-10}
			& IN-LT & IN-100 & IN-200 & IN-LT & IN-100 & IN-200 & IN-LT & IN-100 & IN-200 \\
			\hline
			Softmax & 37.8 & 81.7 & 79.7 & 53.3 & 79.7 & 78.4 & 0 & 0 & 0 \\
			CPN  & 37.1 & {86.1} & 82.1 & 54.5 & 82.3 & 79.5 & 2M & 0.2M & 0.4M \\
			RPL  & 39.0 & 81.8$^{*}$ & 80.7$^{*}$ & 55.1 & 81.2$^{*}$ & 80.2$^{*}$ & 2M & 0.2M & 0.4M \\
			RPL++  & 39.7 & - & - & 55.2 & - & - & 4M & - & - \\
			\hline
			PMAL & \textbf{42.9} & \textbf{86.2} & \textbf{84.1} & \textbf{71.7} & \textbf{94.9} & \textbf{93.9} & 0 & 0 & 0 \\
			\hline
		\end{tabular}
	}
\end{table*}
The same metrics (ACC and AUROC) are used for evaluation.
See Table \ref{sota_imagenet}, our method improves close set ACC by 3.2\% and 2\% on ImageNet-LT and ImageNet-200.
Moreover, open set AUROC is significantly enhanced by 16.5\%, 12.6\% and 13.7\% compared to state-of-the-art.
Such performance gains are much more evident than the improvements on small-scale datasets, reflecting that our method possesses larger advantage in more challenging large-scale tasks.
It can be observed that the parameter number (apart from the adopted same backbone) of previous methods increases along with known class number, which exceeds a non-negligible cost 2M on ImageNet-LT. 
The increased prototype parameters may aggravate the difficulty of model training, thus deteriorate their performance.
Instead, since PMAL brings no parameters for prototypes, its performance is invariant to the number of known classes, which explains the promotion on more complicated datasets.

{Besides, the huge promotion on \textit{long-tailed} dataset shows PMAL can better handle the problem than the existing. Previous ones tend to learn unreliable prototypes for those few-sample classes, due to unbalanced training of various classes.
	While PMAL directly mines prototypes from training data, thus produces stable prototypes with fewer samples.}

\begin{table*}[t]
	\centering
	\caption{Ablations of each module on TinyImageNet.}
	\label{tab_pl_compare}
	\scalebox{0.9}{
		\begin{tabular}{c|c|cccccc}
			\hline
			\multicolumn{2}{c|}{Components} & (a) & (b) & (c) & (d) & (e) & (f) \\
			\hline
			\multirow{2}*{PM} & High-Quality & \checkmark & & \checkmark & & \checkmark & \checkmark \\
			\cline{2-8}
			& Diversity & & \checkmark & & \checkmark & \checkmark & \checkmark \\
			\hline
			EO & Point-to-Set & & & \checkmark & \checkmark &  &\checkmark \\
			\hline
			\multicolumn{2}{c|}{AUROC}  & 80.3 & 78.1 & 81.6 & 80.2 &  81.9 &\textbf{83.1}\\
			\hline
		\end{tabular}
	}
\end{table*}

\subsection{Detailed Analysis}

\begin{figure}[t]
	\centering
	\includegraphics[width=1.\linewidth]{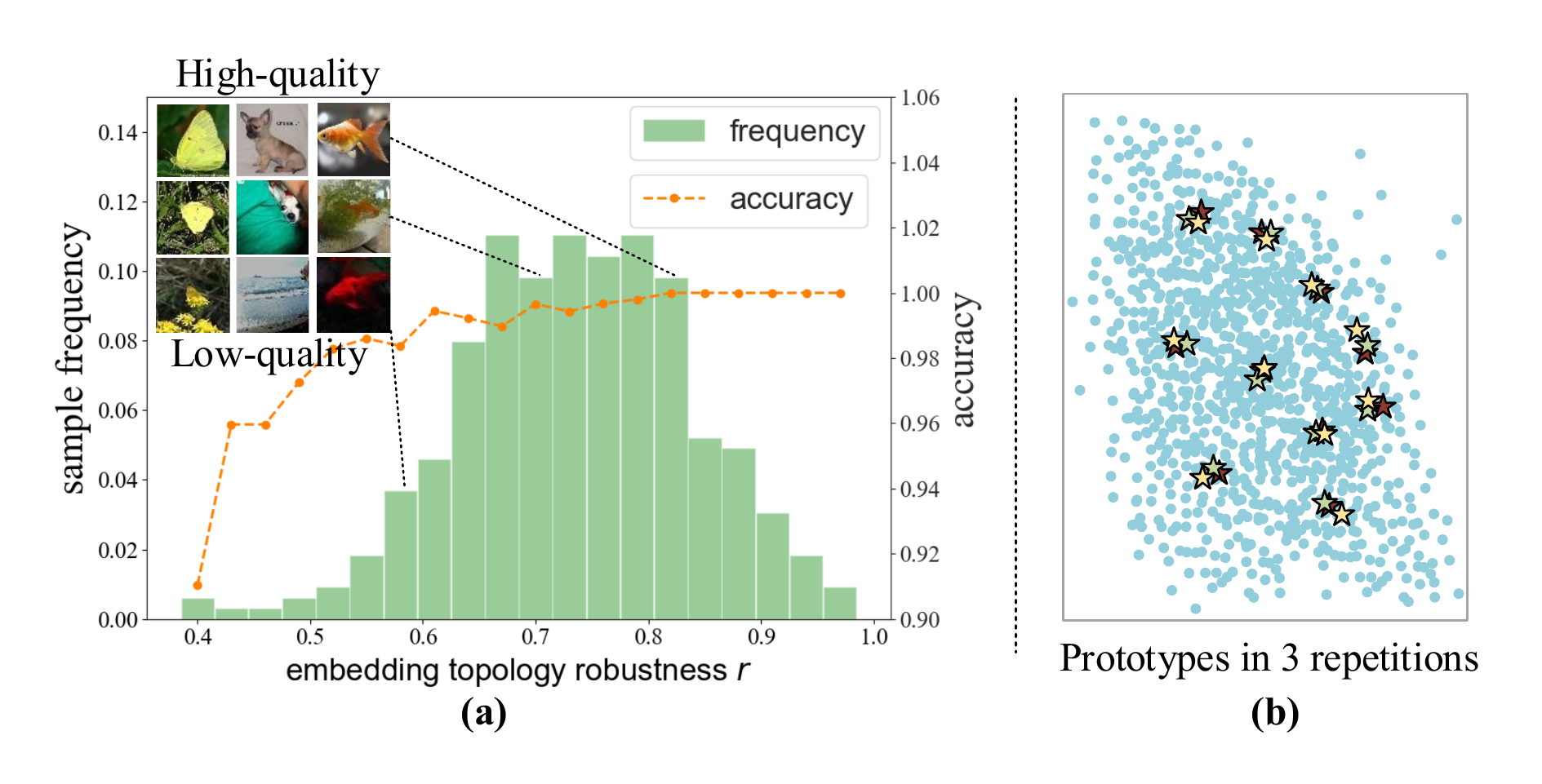}
	\caption{(a) Distribution of $r$; (b) Prototypes in embedding space (visualized by T-SNE) under 3 repetitions. Star in different color denotes prototypes in different repetition.}
	\label{fig-robust-distri}
\end{figure}

\begin{figure}[t]
	\centering
	\includegraphics[width=1.0\linewidth]{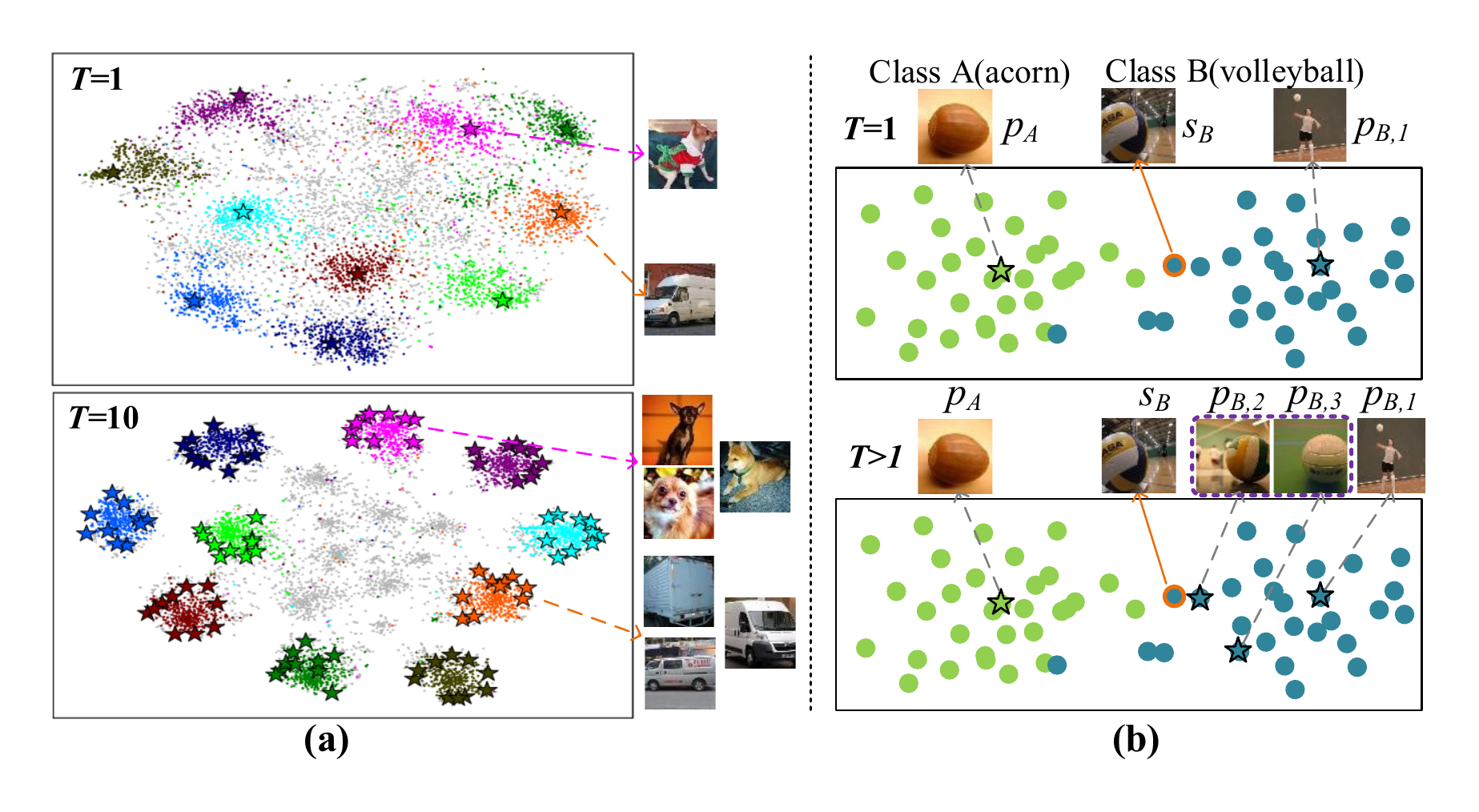}
	\caption{(a)Embedding space (visualized by 2-D T-SNE) of different prototype number; (b)Illustration for the advantage of multiple prototypes. Classes are distinguished by color, star denotes prototype and circle denotes other samples.}
	\label{fig-diver-advan}
\end{figure} 

\begin{figure*}[t]
	\centering
	\includegraphics[width=1.\linewidth]{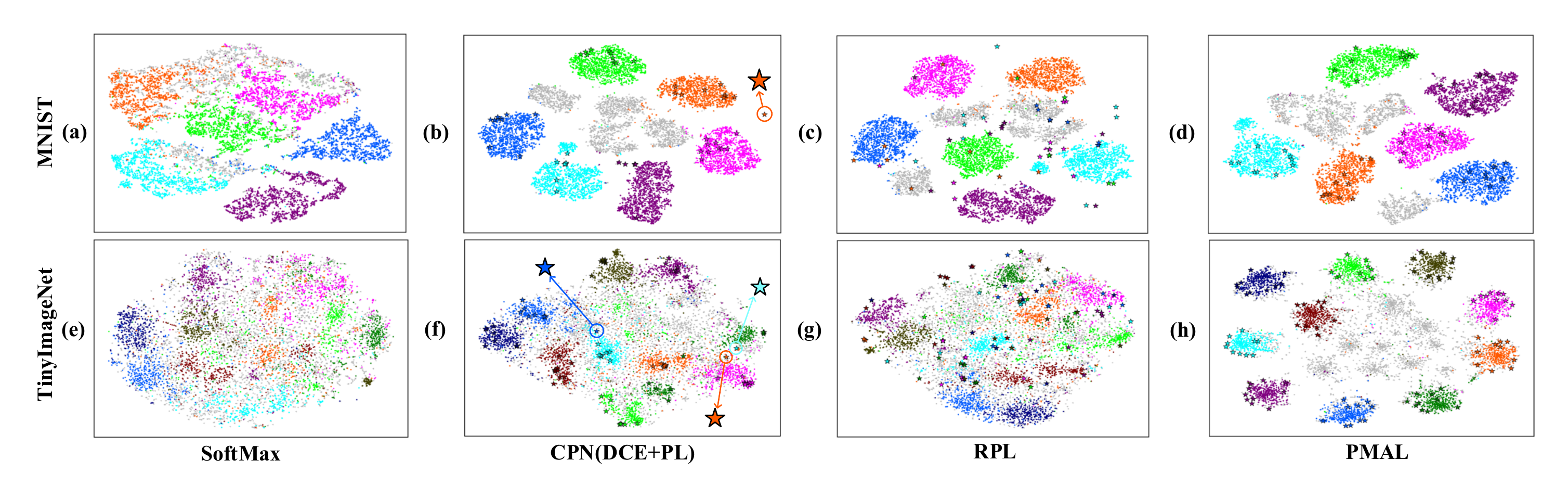}
	\caption{Learned embedding visualized by 2-D T-SNE. CPN and RPL are optimized to achieve reported results. We visualize 10 from the 20 known classes on TINY for better clarity. Each color denotes different classes and `gray' denotes unknowns, prototypes are marked as stars. Better viewed by zooming in.}
	\label{fig-vis-embed-comparison}
\end{figure*}

\begin{table}[t]
	\centering
	\caption{Comparisons with other methods on the \textit{quality} and \textit{diversity} property.}
	\label{tab_unsuper_comparison}
	\scalebox{0.9}{
		\begin{tabular}{ccc}
			\hline
			{Method} & {ACC} & {AUROC}                   \\
			\hline
			\specialrule{0em}{1.pt}{1.pt}
			(a)Probability & 81.9 & 79.3 \\
			\specialrule{0em}{1.pt}{1.pt}
			(b)Deep Ensembles & 82.3 & 80.5 \\
			\specialrule{0em}{1.pt}{1.pt}
			(c)MC-dropout & 81.6 & 78.8 \\
			\hline
			\specialrule{0em}{1.pt}{1.pt}
			(a)Randomization & 81.5 & 79.1 \\
			\specialrule{0em}{1.pt}{1.pt}
			(b)Clustering & 81.8 & 79.6 \\
			\hline
			\specialrule{0em}{1pt}{1pt}
			{Ours} & \textbf{84.7} & \textbf{83.1} \\
			\hline
		\end{tabular}
	}
	\hspace{0.15cm}
	\centering
	\caption{AUROC under different hyper-parameters, including $T$,$\epsilon$,$U$ and $\delta$.}
	\label{tab_hyper_params}
	\scalebox{0.9}{
		\begin{tabular}{c|ccccc}
			\hline
			$T$ & 1 & 5 & 10 & 20 & 30\\
			AUROC & 79.9 & 81.1 & 83.1 & 82.6 & 83.1 \\
			\hline
			$\epsilon$ & 0.1 & 0.3 & 0.5 & 0.7 & 0.9 \\
			AUROC & 73.6 & 78.1 & 82.6 & 83.1  & 81.2 \\
			\hline
			$U$ & 2 & 3 & 4 & 5 & 6 \\
			AUROC & 83.1 & 83.3 & 83.2 & 83.0 & 83.3 \\
			\hline
			$\delta$ & 0.1 & 0.3 & 0.5 & 0.8 & 1. \\
			AUROC & 80.9 & 82.8 & 83.1 & 82.1 & 80.5 \\
			\hline
		\end{tabular}
	}
\end{table}
\subsubsection{Ablation of Each Component in PMAL.}
As shown in Table \ref{tab_pl_compare}, we perform each experiment of the proposed components for ablation study,
including the components regarding the two properties, \ie, \textit{high-quality} and \textit{diversity} (see Sec. 4), and the embedding learning procedure using the \textit{point-to-set} distance metric (see Sec. 5.1).
\begin{itemize}
	\item In prototype mining (PM) phase, both \textit{high-quality} and \textit{diversity} matters in prototypes, where \textit{high-quality} is the more crucial factor validated by our experiments. Jointly combining the two further boosts the performance.
	\item In embedding optimization (EO) phase, no `$\checkmark$' denotes the commonly adopted way: computing the distance from sample $x_i$ to its nearest prototype. Obviously, our proposed \textit{point-to-set} metric has 1.2\% gain, comparing (a)/(c),(b)/(d) or (e)/(f). 
\end{itemize}

\subsubsection{Effect of Embedding Topology Robustness.}
{\textbf{(1) Distribution of $r(\cdot)$}}: We further analyze the distribution of $r(\cdot)$ on TinyImageNet in Figure \ref{fig-robust-distri}(a). It shows different samples correspond to various robustness,
which decreases along with quality degradation caused by occlusion or background interference \etc \
Besides, we summarize classification accuracy in each interval of $r(\cdot)$, which shows a monotonously ascending trend. It means high-quality samples have larger probabilities to be correctly identified.
{\textbf{(2) Comparison with other methods}}: Furthermore, we compare with mainstream methods to prove the advantage of \textit{embedding topology robustness} $r(\cdot)$ for selecting high-quality samples on TinyImageNet:
(a) \textit{Probability}: samples with the highest predicted probability
from each class are chosen as prototypes;
(b) \textit{Deep Ensembles} \cite{Balaji2017Simple}: samples with the lowest probability variance between two models are used as prototypes;
(c) \textit{MC-dropout} \cite{Yarin2016Dropout}: data uncertainty is modeled by MC-dropout with ratio 0.5 and samples with the lowest probability variance are selected as prototypes.
For a fair comparison, we fix $\epsilon$ to 0.7 and replace $r(\cdot)$ with above `probability' or `probability variance', and then adopt the same diversity strategy to produce final prototypes.
Table \ref{tab_unsuper_comparison} verifies proposed \textit{embedding topology robustness} is better than other widely-used data uncertainty modeling methods.


\subsubsection{Effect of Diversity-based Filtering.}
{\textbf{(1) Number of prototypes:}} We vary prototype number $T$ to quantify the effect of diversity on TinyImageNet. As in Table \ref{tab_hyper_params}, diverse prototypes lead to much better performance than a single prototype, we obtain the best result when $T$=10. 
When $T$ increases over 10, performance gradually stabilized. 
To interpret the advantage of diversity, we visualize embedding space when $T$=1 or $T$=10 in Figure \ref{fig-diver-advan}(a). Evidently, the chosen 10 prototypes appear in diverse visual looking and their embeddings are located at separate positions. Moreover, using 10 prototypes learns more compact intra-class embedding regions, where unknown sample points are farther from the embedding region of known classes.
The reason can be illustrated by Figure \ref{fig-diver-advan}(b): when $T$=1, sample $s_B$ from class B looks even more like the prototype $p_A$ of class A than the prototype $p_{B,1}$ of class B, but its embedding will still be forced to be closer to $p_{B,1}$ than $p_A$, which is hard to be optimized. Instead when $T$$>$1, embedding of $s_B$ is mainly pulled closer to $p_{B,2}$ and $p_{B,3}$ with similar appearance and adjacent embedding, which eases training difficulty. Hence it learns more compact intra-class embedding.
{\textbf{(2) Comparison with other methods}}:
We compare with 2 strategies to select multiple ($T$=10) prototypes from same candidates:
(a) \textit{Randomization}: prototypes are randomly selected from candidates.
(b) \textit{Clustering}: K-Means clustering is used and samples whose embedding nearest to cluster centers are used as prototypes.
Table \ref{tab_unsuper_comparison} shows the obvious advantage using our diversity-based method above randomization and clustering.
\subsubsection{Fluctuation of Prototypes.}
The selected prototypes are affected by pre-trained embedding model $M^1$ and $M^2$. We repeat the mining process for 3 times using different embedding models, and give the results of a class from TinyImageNet in Figure \ref{fig-robust-distri}(b). We observe prototypes chosen in different repetitions only fluctuate very slightly in embedding space, which validates the stability of prototype mining.

\subsubsection{Visualization of Embedding Space.} 
We compare learned embedding space of PL methods in Figure \ref{fig-vis-embed-comparison}. On the easier MNIST dataset, compared to the naive `SoftMax',
the latter three can enlarge the distance among known classes.
But on the more complex TinyImageNet, our method pushes away the embedding region of known classes to a much larger extent compared to CPN and RPL, so the overlap between known and unknowns is much more evidently reduced.

For CPN, we observe undesired prototypes in the circle of Figure \ref{fig-vis-embed-comparison}(f), arising from the unstable prototype learning fooled by low-quality samples.
For RPL,
known and unknown samples also can not be well separated by referring to learned prototypes in Figure \ref{fig-vis-embed-comparison}(g).
This reveals existed methods endanger from learning sub-optimal prototypes or embedding space especially in complicated tasks.
Instead, PMAL is capable of mining trustworthy prototypes and optimizing satisfying embeddings in more challenging tasks.

\subsubsection{Hyper-parameters.}
(1) \textbf{Threshold $\epsilon$}. $\epsilon$ controls the balance of prototype quality and diversity. A larger $\epsilon$ implies more strict condition to ensure quality, but less samples will be elected as candidates, thus diversity among candidates is reduced. A smaller $\epsilon$ is on the contrary. As in Table \ref{tab_hyper_params}, setting $\epsilon$ too small or large both results in AUROC declined. We find $\epsilon$ in the range [0.6, 0.8] leads to stable performance.
(2) \textbf{Initial model number $U$}. Optionally we can adopt more than 2 models to compute $r(\cdot)$, that is, we first compute $r(\cdot)$ with Equation \ref{equa-robustness} between two arbitrary models then average the results. AUROC remains similar as we increase $U$, shown in Table \ref{tab_hyper_params}, which implies 2 models are already sufficient to extract effective prototypes. Note that we adopt 2 models for prototype mining, but only one model is employed for prototype learning, thus no extra model parameters are added during inference.
(3) \textbf{Margin $\delta$ and $\lambda_{p}$}. $\delta$ decides the separability of different embedding regions. When $\delta$ is too small, different regions can not be well separated. But if $\delta$ becomes too large, $\mathcal{L}_{p}$ will grow to a large value overwhelming $\mathcal{L}_{cls}$, causing $\mathcal{L}_{cls}$ hard to converge.
See Table \ref{tab_hyper_params}, a value between 0.3 and 0.5 for $\delta$ can produce better AUROC results.
For loss weight $\lambda_{p}$, when we vary it among 0.5, 0.8 and 1, the resulted AUROC are 82.6, 82.8 and 82.9, which implies PMAL is not very sensitive to $\lambda_{p}$. For all the validations in various tasks, we adopt the universal hyper-parameter setting: $T$=10, $\epsilon$=0.7, $U$=2, $\delta$=0.5 and $\lambda_{p}$=1.

\section{Conclusion}
This paper proposes a novel prototype mining and learning algorithm.
It directly discovers high-quality and diversified prototype sets from training samples.
Then based on generated prototypes, the OSR model can focus on optimizing a better embedding space in which known and unknown classes are separated.
Extensive experiments on various benchmarks show that our method outperforms the state-of-the-art approaches.
In future work, we will explore our prototype mining mechanism in broader tasks other than OSR.

\bibliography{aaai22}

\end{document}